## Adaptive Image Restoration for Video Surveillance: A Real-Time Approach

[1*]Muhammad Awais Amin, [2]Adama Ilboudo, [3]Abdul Samad bin Shahid, [4]Amjad Ali, [5]Waqas Haider Khan Bangyal

**Article Details**

**Muhammad Awais Amin**
Data Science Consultant, Datamatics Technologies, Gulberg Greens, Islamabad, Pakistan. awais2815@gmail.com

**Adama Ilboudo**
Data Scientist, SaH Analytics International, Abidjan, Ivory Coast & International Data Science Institute, INPHB, Yamoussoukro, Ivory Coast. adama.ilboudo21@inphb.ci

**Abdul Samad bin Shahid**
Business Intelligence Consultant, Datamatics Technologies, Gulberg Greens, Islamabad samadshahid1993@gmail.com

**Amjad Ali**
MS Electronic Engineering, Karachi Institute of Economics and Technology, Karachi, Pakistan. aliyousafzai17@gmail.com

**Waqas Haider Khan Bangyal**
Associate Professor(CS) and Dean of Sciences-Kohsar University Murree (KUM), Muree, Pakistan. waqas.bangyal@kum.edu.pk

## ABSTRACT

One of the major challenges in the field of computer vision especially for detection, segmentation, recognition, monitoring, and automated solutions, is the quality of images. Image degradation, often caused by factors such as rain, fog, lighting, etc., has a negative impact on automated decision-making.Furthermore, several image restoration solutions exist, including restoration models for single degradation and restoration models for multiple degradations. However, these solutions are not suitable for real-time processing. In this study, the aim was to develop a real-time image restoration solution for video surveillance. To achieve this, using transfer learning with ResNet_50, we developed a model for automatically identifying the types of degradation present in an image to reference the necessary treatment(s) for image restoration. Our solution has the advantage of being flexible and scalable

## INTRODUCTION

Today, thanks to computer vision, we are transitioning from a stage where surveillance cameras were constantly monitored by humans who reported events themselves, to a stage where events are automatically detected and reported by intelligent systems connected to these cameras, paving the way for more efficient surveillance and less prone to human errors. [1], [2]. These systems are becoming increasingly advanced. As examples, we have automatic detection of violence [1], Automatic detection of street fights, fire detection, facial recognition, motion detection, object detection, object segmentation, theft detection in shopping malls, license plate detection, anomaly detection, etc. [3].

Automatic monitoring, while offering numerous advantages in terms of efficiency and reducing dependence on human surveillance, also presents its own set of challenges. One of the major challenges which is indeed a fundamental issue in the field of computer vision or machine learning in general is the quality of data, specifically the quality of images in our case. Image degradation, whether due to environmental conditions such as rain, fog, dust, etc., hardware aging, or other factors, can have a significant impact on the accuracy of event detection [4]. If an image is blurry, grainy, or obscured, automated algorithms may miss important elements, which can lead to the non-detection of suspicious cases or anomalies or result in false alarms or misidentification.

Furthermore, several research efforts are directed towards the theme of "Digital Image Restoration," spanning across various fields such as medical imaging, photography, astronomy, satellite imagery, art, surveillance, and security etc. [5], [6], [7]. Thus, several restoration solutions are proposed. Some solutions are focused on "single degradation," while others address "multiple degradations." The former is designed to handle a specific form of degradation such as blur, noise, or JPEG compression. The latter, on the other hand, can address multiple types of degradation; for instance, it can simultaneously address fog, blur, and raindrop effects.

However, the problem with these methods lies in their adaptation to real-time processing. Indeed, "single degradation" restoration methods not only require prior knowledge of the type of degradation but also lack versatility. As for "multiple degradation" restoration methods, they pose an issue of increased complexity, especially when considering a significant number of degradation types.

This paper follows this structure: after the introduction, next point is the literature review. In this point deals with three subpoints: some types of image degradation, definitions of image restoration and last one is image restoration approaches. The next point is about the methodology where we discuss about the identification of the degradation types from images, image restoration techniques for single degradation and image restoration techniques for multiple degradation. And before the conclusion and perceptive, we present and discuss ours results.

## LITTERATURE REVIEW

## TABLE 1: RESEARCH PAPERS

| Title | Tasks | Evaluation |
|---|---|---|
| **Restormer : Efficient Transformer for High-Resolution Image restoration** | Deraining, deblurring, denoising | PSNR = 32.92 and SSIM = 0.961 (GoPro testing dataset) (deblurring) |
| **SwinIR: Image Restoration Using Swin Transformer** | Super-resolution, denoising, and JPEG artifact removal | PSNR = 34.62 and SSIM = 0.9289 (Set5 testing dataset) (ightweight image SR) |
| **Learning Enriched Features for Real Image Restoration and Enhancement** | Denoising, Super-resolution, enhancing | PSNR = 39.72 and SSIM = 0.959 (SIDD dataset) (denoising) |
| **Multi-Stage Progressive Image Restoration** | Deraining, deblurring, denoising | PSNR = 39.80 and SSIM = 0.954 (DND dataset) (denoising) |
| **DeblurGAN-v2: Deblurring (Orders-of-Magnitude) Faster and Better** | Restoring sharp details and fine textures in blurred images, and can handle a wide range of blur types, including motion blur, out-of-focus blur, and Gaussian blur | PSNR = 29.55 and SSIM = 0.934 (GoPro dataset) (debluring) |
| **Uformer : A General U-Shaped Transformer for Image Restoration** | Denoising, deblurring, deraining | PSNR = 26.28 and SSIM = 0.891 (DPD test dataset) (defocus blur removal) |
| **GridDehazeNet :** | Dehazing | PSNR = 32.16 SSIM |
| **Attention-based multiscale network for image dehazing** | | = 0.98 (SOTS INDOOR) (dehazing) |
| **EPDN : Enhanced pix2pix dehazing network** | Dehazing | PSNR = 25.06 and SSIM = 0.92 (SOTS INDOOR) (dehazing) |
| **Unsupervised Single Image Deraining with Self-supervised Constraints** | Deraining | PSNR = 37.28 and SSIM = 0.96 (Rain12 dataset) (deraining) |
| **Unsupervised classspecific deblurring** | Deblurring | PSNR = 21.92 and SSIM = 0.89 (text image dataset.) (Deblurring) |
| **Unsupervised domain-specific deblurring via disentangled representations** | Deblurring | PSNR = 20.81 and SSIM = 0.65 (CelebA dataset.) (Deblurring) |
| **GCDN : Deep graph-convolutional image denoising** | Denoising | PSNR = 41.48 and SSIM = 0.9697 (CelebA dataset.) (Deblurring) |
| **KBNet : Kernel Basis Network for Image Restoration** | Image denoising, deraining, and deblurring | PSNR = 34.19 and SSIM = 0.944 (Test2800 dataset.) (Deraining) |
| **Pre-Trained Image Processing Transformer** | Denoising, super-resolution and deraining | PSNR = 41.62 and SSIM = 0.988 (Rain dataset.) (Deraining) |
| **Satellite image inpainting with deep generative** | Image inpainting, denoising, and image degradation recovery | absolute deviations error of 1 = 0.33, mean squared error |

| adversarial neural | loss of $\mathcal{L}_2 = 0.15$ |
|---|---|
| networks | |

Image restoration has been a longstanding research topic in digital image processing, dating back to the last century, and remains an active area of study in recent years [8].

Image restoration is a process aimed at recovering the original or non-degraded image from a degraded version. [9]. This process involves several tasks, including noise reduction, which aims to eliminate or reduce unwanted artifacts present in the degraded image; dehazing, which reduces the effects of fog or haze; super-resolution, which increases resolution; deblurring, which corrects the blur effect; and deraining, which reduces the effects of rain in an image. [8], [10], [11].

Deep learning techniques, driven by convolutional neural networks, have garnered significant attention in nearly every domain of image processing, including image classification.[8], medical image processing [12], surveillance videos [13], Incident detection, satellite image processing [14], etc. However, image restoration is a fundamental and challenging subject that plays a significant role in the processing, understanding, and representation of images. [8].

Image restoration methods can be classified into two main categories: traditional methods and learning-based methods [15].

## TRADITIONAL METHODS

Traditional methods typically employ an iterative approach to estimate missing information in a corrupted image. [16]. They rely on constructing a mathematical model describing the characteristics of the image to be restored and the corruption process. This model is then used to estimate the missing values in the corrupted image. These methods can be based on statistical models, geometric models, or physical models. [17]. They often require prior knowledge of the corruption process and assumptions about the characteristics of the image. This can make these methods more complex and require intensive computation. Several studies have been conducted using this approach, such as "Fast image super-resolution via local regression." [18], « Seeing through water » [19], « Anchored Neighborhood Regression for Fast Example-Based Super-Resolution » [20].

## LEARNING-BASED METHODS

Learning-based methods use deep neural networks to learn directly from pairs of corrupted and restored images. These methods are often trained on large databases of images to learn how to reconstruct missing or damaged details in an image. Neural networks learn from the information contained in the training data and are capable of generalizing to restore images they have never seen before. These methods have the advantage of not requiring explicit knowledge of the corruption process and can be more flexible in adapting to different types of corruptions.

Learning-based methods can be distinguished into two types of approaches: single degradation image restoration and multiple degradation image restoration. [21].

**IMAGE RESTORATION: SINGLE EGRADATION**

In this approach, models are indeed designed for a single restoration task: either blur correction, noise reduction, raindrop removal, etc. Several studies are based on this approach. For the dehazing task, for example, we have among many others : AECR-Net[22] with CNN architecture, EPDN[23] with GAN architecture, etc., for denoising task, we have : GCDN[24], SADNet[25] with CNN architecture, Adversarial Distortion Learning for Medical Image Denoising using GAN architecture [12], etc., for *deblurring task*, we have : unsupervised classspecific deblurring[26], Unsupervised domain-specific deblurring via disentangled representations[27], etc. This approach has the advantage of being simple (faster execution time) and effective (significantly corrects the specific degradation of the image) [28].

However, using these methods or models implies prior knowledge of the nature of the image degradation. And for each different type of degradation, specific learning is required. However, in reality, with video surveillance, satellite images, old photos, the nature of degradation can vary from one image to another, and an image may suffer from more complex degradations, combining different types of degradation.

**IMAGE RESTORATION: MULTIPLE DEGRADATIONS**

This approach, also known as all-in-one solutions, allows for addressing multiple image restoration tasks without the need for additional training. It aims to simplify the restoration process by avoiding the necessity of treating each type of degradation separately. Pairs of images (degraded, non-degraded) from different degradation classes are used for training so that the model can be capable of restoring images with various types of degradation. To mention just a few: MemNet[29], which is a model based on CNN architecture, enables simultaneous processing of restoration tasks such as denoising, super-resolution, and JPEG deblocking, CVPR-AirNet[21], which addresses denoising, deraining, and dehazing tasks, is also based on CNN architecture. This approach offers a significant advantage in that prior knowledge of degradation is no longer necessary for restoration. However, this approach is more complex in terms of execution time, thus becoming costly for real-time processing.

Image restoration is a subject that has attracted the attention of several researchers. Some have approached it with the traditional approach, while others have used the learning-based approach. The learning-based approach is recommended for complex degradations. It can also be apprehended in two ways: restoration for single degradation types and restoration for multiple degradation types. The former is used for a specific type of degradation and generally has a quick execution time, but it requires prior knowledge of the degradation nature. The latter, on the other hand, is capable of handling multiple types of degradation, thus prior knowledge of the degradation is not always necessary. However, it is generally more complex and requires significant time for execution, making it less practical for real-time restoration processes.

## METHODOLOGY

Real-time image restoration involves versatility regarding the type of degradation and speed in processing. To this end, we have defined an approach that takes these requirements into account. The diagram below outlines our approach in broad strokes.

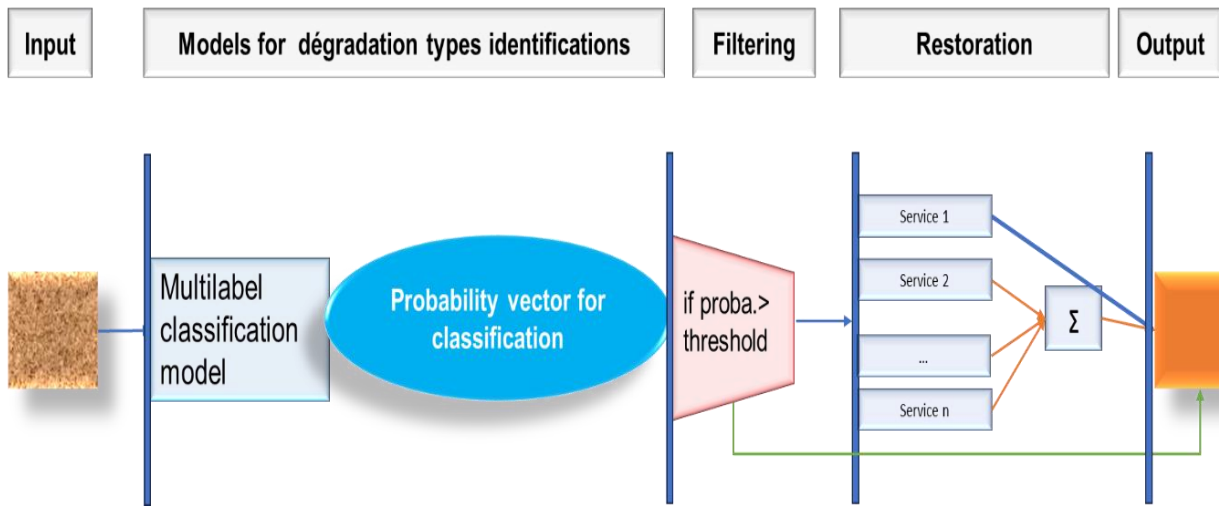

**FIGURE 2: CONCEPTUAL DIAGRAM**

## AUTOMATIC IDENTIFICATION OF DEGRADATION TYPES

In this section, we will address the type of study conducted, the resources used, and the modeling.

## TYPE OF STUDY

Automatic identification of degradation type is a classification problem in the field of machine learning. To benefit from the knowledge of certain existing models, we have opted for transfer learning.

RESSOURCES

DATA

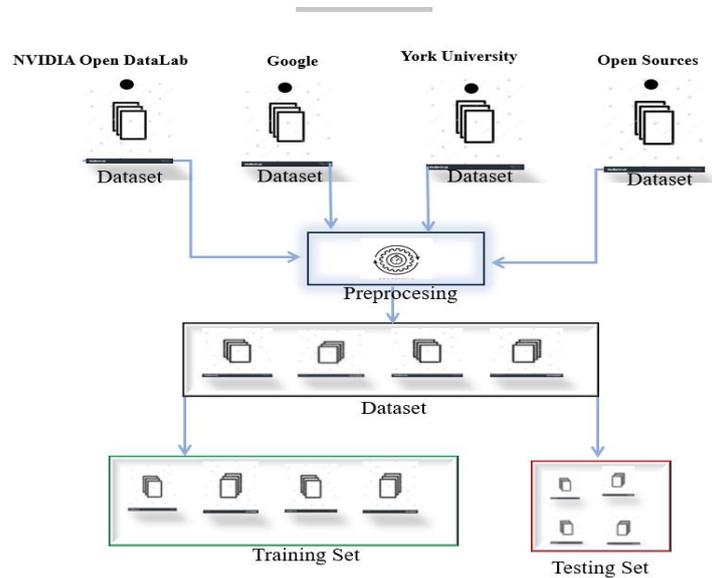

**FIGURE 3: DATAFLOW**

We have a total of 5,187 images in seven classes (Denoising, Dehazing_Indoor, Dehazing_Outdoor, Deblurring, Deraining, Enhancement, Super_resolution). Our images are sourced from GOOGLE, NVIDIA OPEN DATALAB, YORK UNIVERSITY, and other open sources. The name of each class corresponds to the name of the restoration service required. For example, deblurring means a service for reducing or eliminating blur is needed. After retrieving each dataset, data preprocessing operations such as resizing, and normalization were applied.

Our dataset was split into two following a stratified sampling plan (the strata being the types of degradation) with equal proportions and a mode of random sampling without replacement within each stratum. Thus, one dataset was used for training with 4,129 images, and another dataset was used for testing with 1,058 images. The Table below presents the distribution of our dataset according to the type of degradations (tasks to be done) and the set (train and test).

**TABLE 2: DATASET**

| TasKS[source] | Train Set | Test Set | Total |
|---|---|---|---|
| **Deblurring** [30], [31] | 1076 | 275 | **1351** |
| **Dehazing_indoor** [32] | 400 | 105 | **505** |
| **Dehazing_outdoor** [32] | 400 | 105 | **505** |
| **Denoising** [33] | 525 | 130 | **655** |
| **Deraining** [34] | 688 | 178 | **866** |
| **Enhancement** [35], [36] | 400 | 100 | **500** |
| **Super_resolution** [37] | 640 | 165 | **805** |
| **Total** | **4129** | **1058** | **5187** |

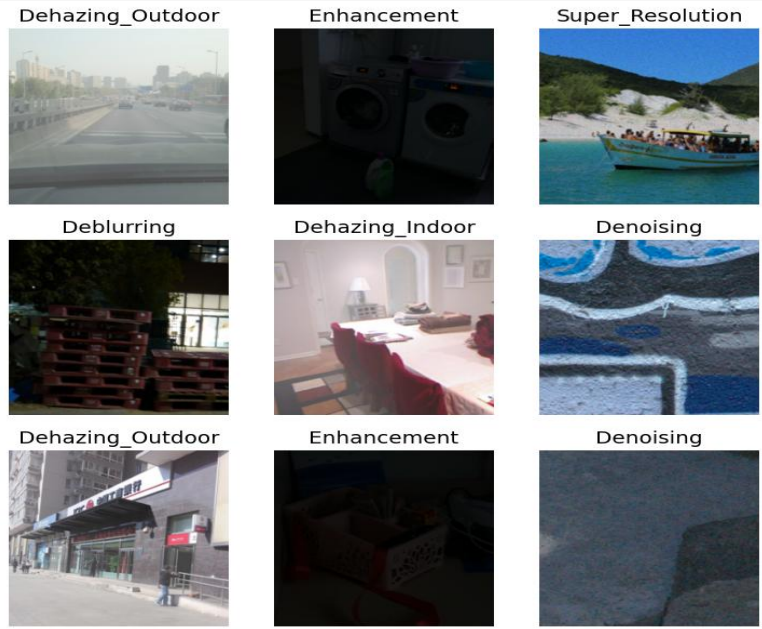

FIGURE 4: SOME PREVIEWS OF IMAGES FROM THE DATASET

## COMPUTING MACHINE

The computing machine we worked with is an MSI GL65 Leopard, running Windows 11 Home as the operating system, equipped with a 512GB SSD for storage, 16GB of RAM, 8GB of GPU, and an Intel(R) Core(TM) i7-10750H CPU @ 2.60GHz, 2592 MHz, 6 cores, 12 logical processors.

## MODELING

ResNet is a convolutional neural network (CNN) architecture introduced in 2015. ResNet outperformed other models such as VGG, GoogleLeNet, and BN-Inception in the ImageNet Large Scale Visual Recognition Challenge (ILSVRC) in 2015, achieving a top-5 error rate of only 3.57%.[38]. Note that the ImageNet dataset contains over fourteen million images. Therefore, to leverage the knowledge gained by ResNet on the

ImageNet dataset and apply it to our own classification problem, we applied transfer learning on variants of ResNet: ResNet_34, ResNet_50, Resnext50_32x4d, and EfficientNet_b2.

## CLASSIFICATION WITH RESNET

A good image classification often results from effective feature extraction. In this regard,

ResNet is considered to have an architecture conducive to feature extraction. As shown by some researchers, it outperforms other models in image classification. ResNet offers several advantages. Indeed, to address the problem of "Vanishing gradient" during the training of deep neural networks, ResNet utilizes residual connections or "skip connections" or "shortcuts" in residual blocks.

**TABLE 4 : ARCHITECTURE RESNET**

| layer name | output size | 18-layer | 34-layer | 50-layer | 101-layer | 152-layer |
|---|---|---|---|---|---|---|
| conv1 | 112×112 | 7×7, 64, stride 2 | | | | |
| conv2_x | 56×56 | 3×3 max pool, stride 2 | | | | |
| | | $\begin{bmatrix} 3\times3, 64 \\ 3\times3, 64 \end{bmatrix} \times 2$ | $\begin{bmatrix} 3\times3, 64 \\ 3\times3, 64 \end{bmatrix} \times 3$ | $\begin{bmatrix} 1\times1, 64 \\ 3\times3, 64 \\ 1\times1, 256 \end{bmatrix} \times 3$ | $\begin{bmatrix} 1\times1, 64 \\ 3\times3, 64 \\ 1\times1, 256 \end{bmatrix} \times 3$ | $\begin{bmatrix} 1\times1, 64 \\ 3\times3, 64 \\ 1\times1, 256 \end{bmatrix} \times 3$ |
| conv3_x | 28×28 | $\begin{bmatrix} 3\times3, 128 \\ 3\times3, 128 \end{bmatrix} \times 2$ | $\begin{bmatrix} 3\times3, 128 \\ 3\times3, 128 \end{bmatrix} \times 4$ | $\begin{bmatrix} 1\times1, 128 \\ 3\times3, 128 \\ 1\times1, 512 \end{bmatrix} \times 4$ | $\begin{bmatrix} 1\times1, 128 \\ 3\times3, 128 \\ 1\times1, 512 \end{bmatrix} \times 4$ | $\begin{bmatrix} 1\times1, 128 \\ 3\times3, 128 \\ 1\times1, 512 \end{bmatrix} \times 8$ |
| conv4_x | 14×14 | $\begin{bmatrix} 3\times3, 256 \\ 3\times3, 256 \end{bmatrix} \times 2$ | $\begin{bmatrix} 3\times3, 256 \\ 3\times3, 256 \end{bmatrix} \times 6$ | $\begin{bmatrix} 1\times1, 256 \\ 3\times3, 256 \\ 1\times1, 1024 \end{bmatrix} \times 6$ | $\begin{bmatrix} 1\times1, 256 \\ 3\times3, 256 \\ 1\times1, 1024 \end{bmatrix} \times 23$ | $\begin{bmatrix} 1\times1, 256 \\ 3\times3, 256 \\ 1\times1, 1024 \end{bmatrix} \times 36$ |
| conv5_x | 7×7 | $\begin{bmatrix} 3\times3, 512 \\ 3\times3, 512 \end{bmatrix} \times 2$ | $\begin{bmatrix} 3\times3, 512 \\ 3\times3, 512 \end{bmatrix} \times 3$ | $\begin{bmatrix} 1\times1, 512 \\ 3\times3, 512 \\ 1\times1, 2048 \end{bmatrix} \times 3$ | $\begin{bmatrix} 1\times1, 512 \\ 3\times3, 512 \\ 1\times1, 2048 \end{bmatrix} \times 3$ | $\begin{bmatrix} 1\times1, 512 \\ 3\times3, 512 \\ 1\times1, 2048 \end{bmatrix} \times 3$ |
| | 1×1 | average pool, 1000-d fc, softmax | | | | |
| FLOPs | | $1.8\times10^9$ | $3.6\times10^9$ | $3.8\times10^9$ | $7.6\times10^9$ | $11.3\times10^9$ |

**SOURCE : DEEP RESIDUAL LEARNING FOR IMAGE RECOGNITION[38].**

**RESIDUAL BLOCKS**

In a traditional neural architecture, each layer tries to learn a specific transformation of the input data. However, in ResNets, each residual block learns a "residual" of that transformation rather than the transformation itself.

Mathematically,

Consider $x$ and $y = F(x)$ as an input (image) and a desired transformation, respectively.

Instead of learning $y$ directly, the residual block tries to learn the residual:

$$R(x) = y - x \,.$$

Final output:    $y = R(x) + x$ $\qquad\qquad$ (1)

Each residual block can be represented as follows:

$$y = R(x, \{W_i\}) + x \quad (2)$$

Where :

- $x$ is the input to the residual block.
- $R(x, \{W_i\}),$ represents the transformation applied to $x$ within the residual block. This transformation is parameterized by a set of weights $\{W_i\}$ that are associated with the various layers contained in the block.

- $y$ is the output of the residual block, obtained by adding the residual transformation $R(x, \{W_i\})$ to the original input $x$.

Note that the weights in ResNet are initialized using Stochastic Gradient Descent (SGD) with standard Momentum parameters.

## STOCHASTIC GRADIENT DESCENT WITH MOMENTUM

Stochastic Gradient Descent with Momentum is an optimization method that accelerates convergence and avoids local minima by using a weighted average of past gradients. [39]. Cette méthode ajoute un hyper paramètre supplémentaire, le Momentum, qui est en effet a parameter that controls the inertia of the learning process and allows it to overcome areas of low slopes or saddle points. The weight update formula is as follows:

$$W_{t+1} = W_t - \alpha * V_{t+1} \qquad (3)$$

with:

- $\alpha$ : the learning rate.
- $V_{t+1} = \beta * V_t + (1 - \beta) * g_t$
- $\beta$ is the parameter for momentum and $g_t$ is the gradient computed on a batch or mini batch at iteration t.

The Momentum parameter must be chosen carefully as it can influence the stability and accuracy of the solution. The default value used in the original paper for this parameter is 0.9 [38].

Several other estimators that could enhance learning exist: Adam, SGD, etc.

## PERFORMANCE EVALUATION

There are several criteria for evaluating the performance of a machine learning model. In the case of a classification problem, we can apply metrics such as accuracy, sensitivity, and specificity on the test dataset to assess the actual performance of the model. Note: C is the confusion matrix, P is the number of classes, and $C_{ii}$ represents the diagonal elements of the confusion matrix, i.e., the number of correct predictions for each class.

**ACCURACY:** It measures the rate of correct predictions across all individuals. To calculate accuracy, we use the confusion matrix. The formula to calculate accuracy is as follows:

$$Accuracy = \frac{\sum_{i=1}^{p} C_{ii}}{\sum_{i=1}^{p} \sum_{j=1}^{n} C_{ij}} \qquad (4)$$

**SENSITIVITY:** It measures the proportion of true positives among individuals who actually belong to the positive class. To calculate sensitivity in the case of multiple classes, we use the confusion matrix. The formula to calculate sensitivity for a specific class is as follows:

$$Sensitivity = \frac{C_{ii}}{\Sigma_{j=1}^{p} C_{ij}} \quad (5)$$

**SPECIFICITY:** It measures the proportion of true negatives among individuals who do not belong to the positive class. To calculate specificity in the case of multiple classes, we use the confusion matrix. The formula for calculating specificity for a specific class is as follows:

$$Spécificity = \frac{C_{jj}}{\Sigma_{i=1}^{p} C_{ij}} \quad (6)$$

$C_{jj}$ represents the number of correct predictions, that is, true negatives.

## DEFINITION OF RESTORATION METHODS

In this section, we explore the use of pre-trained models for each image restoration task and define a technique for restoring images affected by multiple degradations.

Several image restoration models exist. However, the choice of models is also a critical task in real-time processes. Such processes require short execution times, while restoration itself demands high-quality output. There is a trade-off to consider, given that the quality of restoration is often positively correlated with the model's complexity, as is the case between complexity and execution time. Additionally, the choice of model was influenced by the data used for training. Indeed, as our target is surveillance camera images, we prioritized models trained on these types of images.

With the exception of super-resolution, for tasks such as denoising, indoor dehazing, outdoor dehazing, deblurring, deraining, and enhancement, we have relied on the Multi-Axis MLP for Image Processing (MAXIM).

The MAXIM model is indeed structured according to a hierarchical architecture similar to UNet and supports long-range interactions enabled by spatially gated Multi-Layer Perceptrons (MLP) [40]. Additionally, as illustrated in the figure below, MAXIM achieves peak performance in terms of restoration quality across various tasks. Moreover, regarding complexity, MAXIM requires fewer or a comparable number of parameters and FLOPs (Floating Point Operations Per Second, a measure of computer performance used to quantify the number of floating-point operations a core, machine, or system can perform in one second) compared to competing models such as MPRNet, HINet, AECR, and UEGAN, etc.

**TABLE 5: COMPARISONS OF MODELS**

| Tâches | Dataset | Modèles | PSNR | Parameters(million) | FLOPs (Giga) |
|--------|---------|---------|------|---------------------|--------------|
| Denoising | SIDD | MPRNet | 39,71 | 15,7 | 1176 |
| | | MIRNet | 39,72 | 31,7 | 1572 |
| | | **MAXIM** | **39,96** | **22,2** | **339** |
| Debluring | GoPro | MPRNet | 32,66 | 20,1 | 1554 |
| | | HINet | 32,71 | 88,7 | 341 |
| | | IPT | 32,58 | 114 | 1188 |
| | | **MAXIM** | **32,86** | **22,2** | **339** |
| Deraining | Rain13k | MSPFN | 30,75 | 21,7 | – |
| | | MPRNet | 32,73 | 3,64 | 297 |
| | | **MAXIM** | **33,24** | **14,1** | **216** |
| Dehazing | Indoor | MSBDN | 33,79 | 31,3 | 83 |
| | | FFA-Net | 36,36 | 4,5 | 576 |
| | | **MAXIM** | **39,72** | **14,1** | **216** |
| Enhancement | LOL | MIRNet | 24,14 | 31,7 | 1572 |
| | | **MAXIM** | **23,43** | **14,1** | **216** |

Source : MAXIM: Multi-Axis MLP for Image Processing [40]

For each task, a specific model is trained with a dataset specific to that task. Several relevant datasets have been used as a training base. As shown in the figure below, these include the Denoising Dataset for Smartphone Cameras (SIDD), photographs taken with cameras by a group of photographers, RESIDE-ITS for indoor residence photos, and RESIDE-OTS for outdoor residence photos etc.

**TABLE 6: DATASET USED BY MAXIM SOLUTION**

| tasks | Dataset | Data size |
|---|---|---|
| Denoising | SIDD | 320 |
| Deblurring | GoPro | 2103 |
| | RealBlur-J | 3758 |
| | RealBlur-R | 3758 |
| | REDS | 24000 |
| Deraining | Rain 14000\|1800\|800\|100L\|1200\|12 | 13712 |
| | RaindDrop | 861 |
| Dehazing_indoor | RESIDE-ITS | 13990 |
| Dehazing_outdoor | RESIDE-OTS | 313950 |
| Enhancement | LOL | 4500 |
| | MIT-Adove FiveK | 485 |

Source : MAXIM: Multi-Axis MLP for Image Processing [40]

For the task of super-resolution, we used the pre-trained Residual Dense Network (RDN) model from the Python ISR library. Table6 shows that the RDN model achieves peak performance in super-resolution compared to other super-resolution models.

**TABLE 7: COMPARISON OF MODELS ON VARIOUS BENCHMARK DATASETS FOR SUPER-RESOLUTION.**

| Dataset | Scale | Bicubic | SRCNN [3] | LapSRN [13] | DRRN [25] | SRDenseNet [31] | MemNet [26] | MDSR [17] | RDN (ours) | RDN+ (ours) |
|---|---|---|---|---|---|---|---|---|---|---|
| Set5 | ×2 | 33.66/0.9299 | 36.66/0.9542 | 37.52/0.9591 | 37.74/0.9591 | -/- | 37.78/0.9597 | 38.11/0.9602 | 38.24/0.9614 | **38.30/0.9616** |
| | ×3 | 30.39/0.8682 | 32.75/0.9090 | 33.82/0.9227 | 34.03/0.9244 | -/- | 34.09/0.9248 | 34.66/0.9280 | 34.71/0.9296 | **34.78/0.9300** |
| | ×4 | 28.42/0.8104 | 30.48/0.8628 | 31.54/0.8855 | 31.68/0.8888 | 32.02/0.8934 | 31.74/0.8893 | 32.50/0.8973 | 32.47/0.8990 | **32.61/0.9003** |
| Set14 | ×2 | 30.24/0.8688 | 32.45/0.9067 | 33.08/0.9130 | 33.23/0.9136 | -/- | 33.28/0.9142 | 33.85/0.9198 | 34.01/0.9212 | **34.10/0.9218** |
| | ×3 | 27.55/0.7742 | 29.30/0.8215 | 29.79/0.8320 | 29.96/0.8349 | -/- | 30.00/0.8350 | 30.44/0.8452 | 30.57/0.8468 | **30.67/0.8482** |
| | ×4 | 26.00/0.7027 | 27.50/0.7513 | 28.19/0.7720 | 28.21/0.7721 | 28.50/0.7782 | 28.26/0.7723 | 28.72/0.7857 | 28.81/0.7871 | **28.92/0.7893** |
| B100 | ×2 | 29.56/0.8431 | 31.36/0.8879 | 31.80/0.8950 | 32.05/0.8973 | -/- | 32.08/0.8978 | 32.29/0.9007 | 32.34/0.9017 | **32.40/0.9022** |
| | ×3 | 27.21/0.7385 | 28.41/0.7863 | 28.82/0.7973 | 28.95/0.8004 | -/- | 28.96/0.8001 | 29.25/0.8091 | 29.26/0.8093 | **29.33/0.8105** |
| | ×4 | 25.96/0.6675 | 26.90/0.7101 | 27.32/0.7280 | 27.38/0.7284 | 27.53/0.7337 | 27.40/0.7281 | 27.72/0.7418 | 27.72/0.7419 | **27.80/0.7434** |
| Urban100 | ×2 | 26.88/0.8403 | 29.50/0.8946 | 30.41/0.9101 | 31.23/0.9188 | -/- | 31.31/0.9195 | 32.84/0.9347 | 32.89/0.9353 | **33.09/0.9368** |
| | ×3 | 24.46/0.7349 | 26.24/0.7989 | 27.07/0.8272 | 27.53/0.8378 | -/- | 27.56/0.8376 | 28.79/0.8655 | 28.80/0.8653 | **29.00/0.8683** |
| | ×4 | 23.14/0.6577 | 24.52/0.7221 | 25.21/0.7553 | 25.44/0.7638 | 26.05/0.7819 | 25.50/0.7630 | 26.67/0.8041 | 26.61/0.8028 | **26.82/0.8069** |
| Manga109 | ×2 | 30.80/0.9339 | 35.60/0.9663 | 37.27/0.9740 | 37.60/0.9736 | -/- | 37.72/0.9740 | 38.96/0.9769 | 39.18/0.9780 | **39.38/0.9784** |
| | ×3 | 26.95/0.8556 | 30.48/0.9117 | 32.19/0.9334 | 32.42/0.9359 | -/- | 32.51/0.9369 | 34.17/0.9473 | 34.13/0.9484 | **34.43/0.9498** |
| | ×4 | 24.89/0.7866 | 27.58/0.8555 | 29.09/0.8893 | 29.18/0.8914 | -/- | 29.42/0.8942 | 31.11/0.9148 | 31.00/0.9151 | **31.39/0.9184** |

**SOURCE : RESIDUAL DENSE NETWORK (SUPER RESOLUTION) [41].**

However, considering that an image might be affected by multiple types of degradation, which method should be chosen in such cases?

## RESTORATION TECHNIQUE FOR MULTIPLE DEGRADATIONS

In real life, several scenarios leading to multiple degradations are possible. For instance, we often witness moments of rain and fog, or foggy conditions compounded by lighting issues, or even fluctuating daylight combined with various atmospheric conditions, which can increase the complexity of the degradations. Moreover, another problem is that in the case of multiple degradations, the different degradations do not affect the image to the same extent. For example, for an image affected by raindrops and blur, the image might be more affected by the blur than by the raindrops. Technically, we are dealing with a multi-label classification problem. In other words, an image can belong to multiple classes simultaneously. In the context of restoration, this means that an image can suffer from several types of degradations.

## ACTIVATION FUNCTION

In the ResNet architecture (see Table3), the activation function used in the last layer is the Softmax function. The use of this function in the multi-class case implies that an image can belong to only one class. Mathematically, the Softmax function is defined as follows:

$$Softmax(x)_i = \frac{e^{x_i}}{\sum_{j=1}^{K} e^{x_j}} \quad (7)$$

where :

- $x$ is the input vector.
- $K$ is the total number of classes.

In a multi-label classification problem, the softmax function is not suitable. Therefore, the sigmoid activation function has been used instead of the softmax function in the ResNet architecture.

Mathematically, the sigmoid function is defined as follows::

$$\sigma(z) = \frac{1}{1+e^{-z}} \quad (8)$$

With $z$ , the logit, i.e., the raw input value.

Suppose there are K classes; for each image, the network produces K logits ($z_1$ , $z_2$ , $z_3$ , ..., $z_K$). The sigmoid function is then applied to each $z_i$ individually: $\sigma(z_1), \sigma(z_2), \sigma(z_3), ..., \sigma(z_K)$. The result is a vector of K probabilities. Each value in the vector represents the probability that the instance belongs to the corresponding class.

What is the purpose of this vector of K probabilities?

## MODEL OF RESTORATION FOR MULTIPLE DEGRADATIONS

For an image or frame, two major cases arise: the image is considered as either non-degraded or with tolerable degradation, and the other case, the image is degraded. When the image is degraded, two scenarios also present themselves: the image suffers from a single degradation or it is affected by multiple degradations. In the case of multiple degradations, a specific treatment model is required. Let $\theta$ be threshold indicating whether an image is affected by a particular degradation or not.

then:

$$Image_i \text{ is: } \begin{cases} \textbf{undamaged } if\ Max(\ \sigma(z_1), \sigma(z_2), \sigma(z_3), ..., \sigma(z_K)) < \theta. \\ \qquad \qquad \qquad \cdot \\ \textbf{damaged }, else. \end{cases}$$

And the image is damaged:

$$Image_i \text{ is degradation is}: \begin{cases} \textbf{\textit{unique}} \, if\ \sum_{i=1}^{K} \mathbf{1}_{\sigma(z_i)\, \geq\, \theta} = 1 \\ \qquad \qquad \cdot \\ \textbf{\textit{Multiple}} \, if\ \sum_{i=1}^{K} \mathbf{1}_{\sigma(z_i)\, \geq\, \theta} > 1 \end{cases} \qquad (9)$$

Restoration in the case of multiple degradations can be represented as follows:

$$f(x, \sigma, \varphi, \theta) = \mu \sum_{i=1}^{K} \mathbf{1}_{\sigma(z_i)\, \geq\, \theta} \sigma(z_i)\varphi_i(x)$$

Where:

$\mu = (\sum_{i=1}^{K} \mathbf{1}_{\sigma(z_i)\, \geq\, \theta} \sigma(z_i)\,)^{-1}$ , the inverse of the sum of the probabilities that image (x) belongs to the degradations considered ($\sigma(z_i) \geq \theta$)

$\sigma = (\sigma(z_1), \sigma(z_2), \sigma(z_3), ..., \sigma(z_K))$ , vectors of the probabilities that x belongs to each of the K types of degradation.

$\varphi = \quad (\varphi_1, \varphi_2, \varphi_3, \cdots, \varphi_K)$ , the K restoration functions for the respective K types of degradation.

Note that the probabilities of belonging are perceived as weights of importance for the restoration functions. For example, if $\sigma(z_1)$ is larger, the restoration $\varphi_1$ will be more significant in the restoration process than the others. In other words, the larger $\sigma(z_i)$ ,the more important $\varphi_i$ is. Conversely, the smaller $\sigma(z_i)$, the less important $\varphi_1$ is.

In summary, our approach first involves diagnosing the image to identify the types of degradations that have affected the quality of the image. This initial study leads to one of the following conclusions: either the image is undegraded, the image is affected by a single type of degradation, or the image is affected by multiple types of degradations. To this end, we have created a multilabel classification model by changing the activation function of the last layer (Softmax) of ResNet to the sigmoid function. In the case of a single degradation, an appropriate restoration service (model) is invoked, and in the case of multiple degradations, we proceed with an aggregation for the different types of restorations, taking into account the importance of the degradation in the image.

## RESULTATS ET DISCUSSION

## IDENTIFICATION OF TYPES OF DEGRADATIONS

The first step involved selecting a model from different versions of ResNet (ResNet_34, ResNet_50, ResNext50_32x4d, and EfficientNet_b2). To do this, these five (5) models were trained on the same dataset (Train Set) and evaluated on the same dataset (Test Set). The choice of model was based not only on performance measured by the "accuracy" metric, but also on execution time on the Test Set. As can be seen in the figure below (figure 7), the ResNext_50 model has the best accuracy (89%), followed by ResNet_50 (88%). In terms of execution time, ResNet_34 is in the lead with 106 seconds, followed by ResNet_50 with 108 seconds. Moreover, when considering efficiency, measured by the ratio of "accuracy" to "execution time," the ResNet_50 model is the most efficient. Consequently, ResNet_50 was chosen for further use.

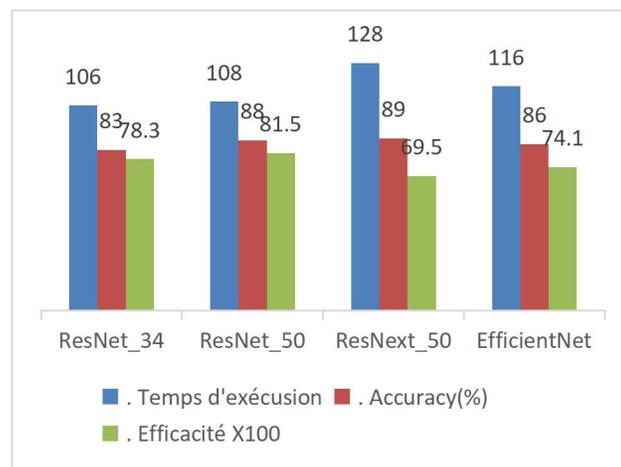

**FIGURE 5: COMPARISON OF MODELS**

Subsequently, the task involved optimizing ResNet_50 to achieve better performance. For this purpose, we based our adjustments on hyperparameters such as the number of iterations (epochs), the number of images per batch (batch size), the learning rate, the optimization function, and evaluated their effects on the validation dataset, which is extracted from the training set at a proportion of 20%.

Initially, keeping other parameters at their default values, the ResNet_50 model was trained over 150 iterations with a batch size of 32 images. However, experimental results showed instability in the model. The batch size was then increased to 64 images, and we observed a decrease in errors on the validation set up to 35 epochs, after which there were no significant improvements. Therefore, we set the number of iterations to 35 and the batch size to 64.

The next steps involved the optimization function and its parameters. We first considered stochastic gradient descent with Momentum. Three values of Momentum (0.9, 0.95, and 0.98) were tested. Additionally, we considered the Adam optimization function with different values of the learning rate. The goal was to find the right combination of optimization functions and either Momentum or learning rate. As can be seen in the Tablebelow, the experimental results show that stochastic gradient descent with a Momentum of 0.95 yields better results than the other tested Momentum values. Indeed, for a Momentum of 0.9 and 0.98, we achieved accuracies of 95% and 92%, respectively.

On the other hand, for the Adam optimization function, three different values were tested (0.001, 0.003, and 0.01). We used the learning rate search method in FastAi (lr_find()), where the suggested rate was 0.003 (see Figure 10), but as this rate might not necessarily be optimal, we bracketed this rate with two others. The model was then evaluated on these values to determine the most optimal parameter. Experimental results showed that the learning rate providing the best performance was 0.001 with an accuracy of 97%.

Finally, we retained the best model for further use, the ResNet_50, with Adam as the optimization function and a learning rate of 0.001, trained over 35 epochs and a batch size of 64.

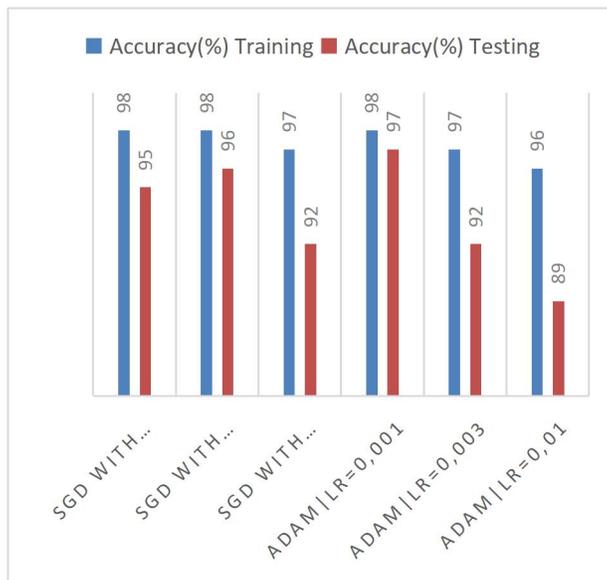

**FIGURE 6: PERFORMANCE SELON LA FONCTION OPTIMISATION**

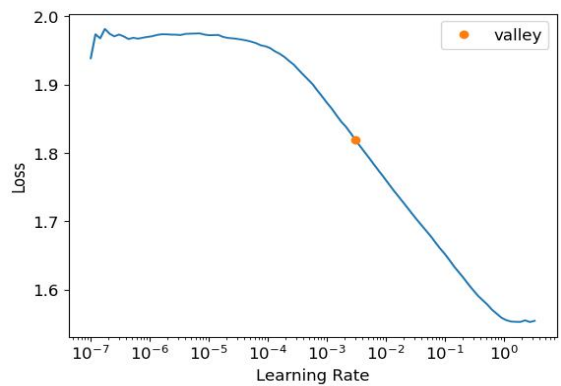

**FIGURE 7: CHOICE OF LEARNING RATE.**

We adopted accuracy as the metric for evaluating the performance of our model. Indeed, we assign the same level of importance to each class. Otherwise, no particular importance is given to any type of error; making a mistake on the "denoising" class is similar to making a mistake on the "deblurring" or "deraining" class, etc. While our data shows little imbalance, a deeper analysis is made on the performance of each class, and this result is presented in Figure 8.

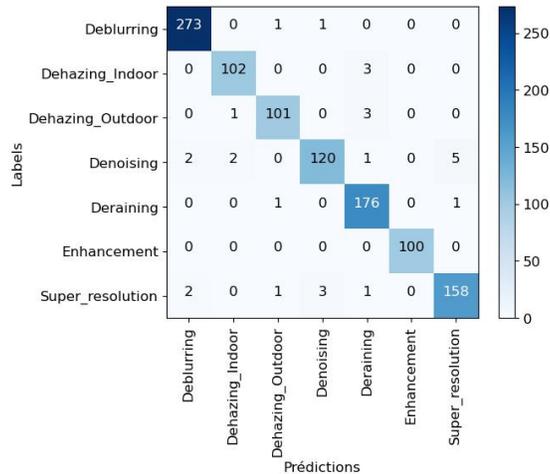

**FIGURE 8: CONFUSION MATRIX.**

No class is penalized by its size, which reassures us about the metric used.

## DEFINITION OF DEGRADATION THRESHOLD

To determine the threshold that defines whether an image is significantly degraded and requires restoration, an analysis is made on the probability of belonging to the classes. Indeed, the threshold was set based on images with no degradation, images with tolerable degradation, and images with significant degradation. It was found that for images without degradation, the probabilities of belonging to the degradation classes are below $0.5$. Whereas for those with tolerable degradation, we have probabilities between $0.5$ and $0.85$. And for images with significant degradation, the probabilities of belonging to the degradation classes were equal to or greater than $0.85$. Consequently, the threshold was set at $0.85$. Thus, an image is considered degraded if it has a probability of belonging to a degradation class equal to or greater than $0.85$, and not degraded otherwise.

## FINAL SOLUTION

The final solution involves coordinating these different methods. To test our solution in a real-time process, we relied on OpenCV, a Python library for computer vision. The goal is to establish an environment capable of connecting to a surveillance camera, either via an internet protocol (IP), or in a local network setting such as WiFi or USB connection, and an environment where it is possible to implement the specific solutions already developed.

The first step is the acquisition of the image, which is captured from the camera in the form of a frame. The next step is diagnosing the frame, attempting to identify the types of degradation affecting the quality of the image. In cases of no degradation or minor degradation, the original frame is output directly, but in cases of significant degradation, the complexity is assessed. If the degradation is singular, a specific restoration service is employed, but in cases of multiple degradations, we aggregate the services of the concerned restorations.

Cases of multiple degradations negatively affect the speed of the process. Indeed, for multiple types of degradation, each type is treated as in the singular case, and then undergoes the aggregation step. However, it is worth noting that these cases of degradation are not very frequent. For example, an image affected by raindrops and fog, especially for an image taken inside residences.

**DISPLAY OF DEGRADED IMAGES AND THEIR RESTORED VERSIONS AFTER AUTOMATIC IDENTIFICATION OF THE REQUIRED RESTORATION TASK.**

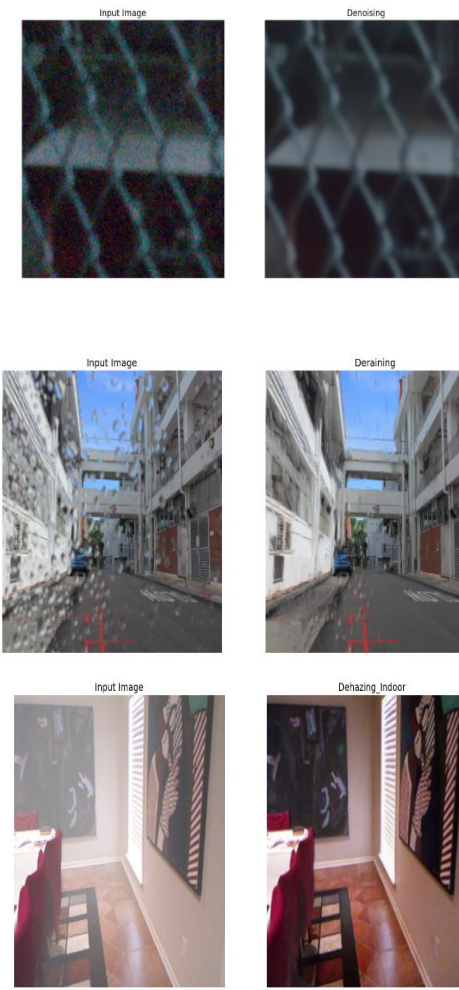

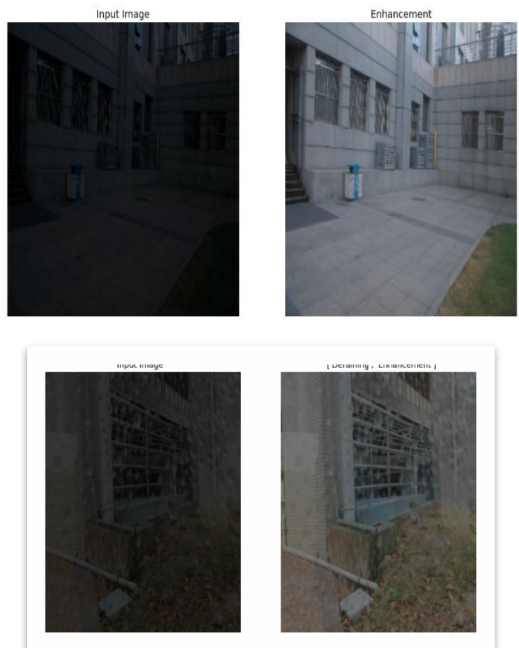

## CONCLUSION AND PERSPECTIVES

Our study aimed to develop a real-time image restoration solution for video surveillance to enhance the robustness of existing computer vision models. An image's quality can be degraded by various environmental and meteorological factors, as well as surveillance hardware. We have limited our focus to seven types of degradation: noise, raindrops, fog effects both inside and outside residences, low resolution, blur, and deterioration such as poor contrast and bad lighting.

To achieve our goal, we used seven pretrained restoration models, each specific to a type of degradation. To benefit from versatility in handling degradations and also improve processing time, we implemented an automatic degradation type identification model. Thanks to this model, each restoration service or model is only called upon if the degradation it addresses is identified.

The identification model was created using transfer learning from the ResNet_50 model, where we modified the last layer by replacing the Softmax activation function with the sigmoid function for multilabel classification. In other words, the model can identify multiple types of degradation present in an image. This model has an accuracy rate of 97%.

The major advantage of our solution is its flexibility and scalability. Indeed, it can be fully or partially updated. The automatic degradation type identification model can be updated to make it more efficient or to add more types of degradation to increase its versatility. The same applies to the image restoration models. Each restoration model can be easily replaced with another more or less complex model depending on resources, and additional models for new types of degradation can be easily added. The solution is easily Adaptable to available resources.

Furthermore, some points for improvement can be noted to guide future studies on the subject. Indeed, training one's models on data specific to video surveillance could help improve restoration accuracy more than using pretrained models, which are more general. Also, exploring a broader range of hyperparameter values than those proposed in this study could increase the chances of finding the best hyperparameters to improve the model's accuracy.

## REFRENCES

[1] S. Vosta et K.-C. Yow, « A CNN-RNN Combined Structure for Real-World Violence Detection in Surveillance Cameras », *Appl. Sci.*, vol. 12, nº 3, Art. nº 3, janv. 2022, doi: 10.3390/app12031021.

[2] N. Sulman, T. Sanocki, D. Goldgof, et R. Kasturi, « How effective is human video surveillance performance? », in *2008 19th International Conference on Pattern Recognition*, Tampa, FL, USA: IEEE, déc. 2008, p. 1‑3. doi: 10.1109/ICPR.2008.4761655.

[3] « Des caméras de surveillance capables d'appeler directement la police », Europe 1. Consulté le: 16 août 2023. [En ligne]. Disponible sur: https://www.europe1.fr/technologies/des-cameras-de-surveillance-capables-dappeler-directement-la-police-3939934

[4] H. Halmaoui, « Restauration d'images par temps de brouillard et de pluie: applications aux aides à la conduite », Université d'Evry-Val d'Essonne, 2012. [En ligne]. Disponible sur: https://theses.hal.science/tel-00830869/file/halmaoui_these.pdf

[5] A. Kaur et G. Dong, « A Complete Review on Image Denoising Techniques for Medical Images », *Neural Process. Lett.*, juill. 2023, doi: 10.1007/s11063-023-11286-1.

[6] I. Goni, A. S. Ahmadu, et Y. M. Malgwi, « Wavelet Transform Technique Applied to Satellite Image Denoising | Electrical Science & Engineering », Nigeria, Modibbo Adama University of Technology, avril 2023. Consulté le: 16 août 2023. [En ligne]. Disponible sur: https://journals.bilpubgroup.com/index.php/ese/article/view/5235

[7] J. Zhang, F. Wang, H. Zhang, et X. Shi, « A Novel CS 2G-starlet denoising method for high noise astronomical image », *Opt. Laser Technol.*, vol. 163, p. 109334, août 2023, doi: 10.1016/j.optlastec.2023.109334.

[8] J. Su, B. Xu, et H. Yin, « A survey of deep learning approaches to image restoration », *Neurocomputing*, vol. 487, p. 46‑ 65, mai 2022, doi: 10.1016/j.neucom.2022.02.046.

[9] M. R. Banham et A. K. Katsaggelos, « Digital Image Restoration », mars 1997. Consulté le: 27 mai 2023. [En ligne]. Disponible sur: https://www.csd.uoc.gr/~hy371/bibliography/ImageRestoration.pdf

[10] B. Kawar, M. Elad, S. Ermon, et J. Song, « Denoising Diffusion Restoration Models ». arXiv, 12 octobre 2022. doi: 10.48550/arXiv.2201.11793.

[11] B. Cai, X. Xu, K. Jia, C. Qing, et D. Tao, « DehazeNet: An End-to-End System for Single Image Haze Removal », *IEEE Trans. Image Process.*, vol. 25, nᵒ 11, p. 5187‑ 5198, nov. 2016, doi: 10.1109/TIP.2016.2598681.

[12] M. Ghahremani, M. Khateri, A. Sierra, et J. Tohka, « Adversarial Distortion Learning for Medical Image Denoising ». arXiv, 29 avril 2022. Consulté le: 28 mai 2023. [En ligne]. Disponible sur: http://arxiv.org/abs/2204.14100

[13] R. F. H. Torres et R. Garcia, « Image Restoration for Blurred License Plates Extracted from Traffic Video Surveillance using Lucy Richardson Algorithm », in *2022 IEEE 14th International Conference on Humanoid, Nanotechnology, Information Technology, Communication and Control, Environment, and Management (HNICEM)*, déc. 2022, p. 1‑ 6. doi: 10.1109/HNICEM57413.2022.10109437.

[14] D. H. Thai, X. Fei, M. T. Le, A. Züfle, et K. Wessels, « Riesz-Quincunx-UNet Variational Auto-Encoder for Satellite Image Denoising ». arXiv, 25 août 2022. Consulté le: 12 mai 2023. [En ligne]. Disponible sur: http://arxiv.org/abs/2208.12810

[15] J. Liang, J. Cao, G. Sun, K. Zhang, L. Van Gool, et R. Timofte, « SwinIR: Image Restoration Using Swin Transformer ». arXiv, 23 août 2021. Consulté le: 9 mai 2023. [En ligne]. Disponible sur: http://arxiv.org/abs/2108.10257

[16] T. GUILLEMOT, « Méthodes et structures non locales pour la restauration d'images et de surfaces 3D », 2021. [En ligne]. Disponible sur: https://www.theses.fr/2014ENST0006.pdf

[17] S. Ishii, S. Lee, H. Urakubo, H. Kume, et H. Kasai, « Generative and discriminative model-based approaches to microscopic image restoration and segmentation », *Microscopy*, vol. 69, nᵒ 2, p. 79‑ 91, avr. 2020, doi: 10.1093/jmicro/dfaa007.

[18] S. Gu, N. Sang, et F. Ma, « Fast image super resolution via local regression », in *Proceedings of the 21st International Conference on Pattern Recognition (ICPR2012)*, nov. 2012, p. 3128‑ 3131.

[19] Y. Tian et S. G. Narasimhan, « Seeing through water: Image restoration using model-based tracking », in *2009 IEEE 12th International Conference on Computer Vision*, sept. 2009, p. 2303‑ 2310. doi: 10.1109/ICCV.2009.5459440.

[20] R. Timofte, V. De, et L. V. Gool, « Anchored Neighborhood Regression for Fast Example-Based Super-Resolution », in *2013 IEEE International Conference on Computer Vision*, Sydney, Australia: IEEE, déc. 2013, p. 1920‑ 1927. doi: 10.1109/ICCV.2013.241.

[21] B. Li, X. Liu, P. Hu, Z. Wu, J. Lv, et X. Peng, « All-in-One Image Restoration for Unknown Corruption ».

[22] H. Wu *et al.*, « Contrastive Learning for Compact Single Image Dehazing ».

[23] Y. Qu, Y. Chen, J. Huang, et Y. Xie, « Enhanced Pix2pix Dehazing Network », in *2019 IEEE/CVF Conference on Computer Vision and Pattern Recognition (CVPR)*, Long Beach, CA, USA: IEEE, juin 2019, p. 8152‑ 8160. doi: 10.1109/CVPR.2019.00835.

[24] D. Valsesia, G. Fracastoro, et E. Magli, « Deep Graph-Convolutional Image Denoising ». arXiv, 19 juillet 2019. Consulté le: 29 mai 2023. [En ligne]. Disponible sur: http://arxiv.org/abs/1907.08448

[25] M. Chang, Q. Li, H. Feng, et Z. Xu, « Spatial-Adaptive Network for Single Image Denoising ». arXiv, 13 juillet 2020. Consulté le: 29 mai 2023. [En ligne]. Disponible sur: http://arxiv.org/abs/2001.10291

[26] N. T. Madam, S. Kumar, et A. N. Rajagopalan, « Unsupervised Class-Specific Deblurring », in *Computer Vision – ECCV 2018*, vol. 11214, V. Ferrari, M. Hebert, C. Sminchisescu, et Y. Weiss, Éd., in Lecture Notes in Computer Science, vol. 11214. , Cham: Springer International Publishing, 2018, p. 358‑ 374. doi: 10.1007/978-3-030-01249-6_22.

[27] B. Lu, J.-C. Chen, et R. Chellappa, « Unsupervised Domain-Specific Deblurring via Disentangled Representations ». arXiv, 5 août 2019. Consulté le: 29 mai 2023. [En ligne]. Disponible sur: http://arxiv.org/abs/1903.01594

[28] M. Tawfik *et al.*, « Comparative Study of Traditional and Deep-Learning Denoising Approaches for Image-Based Petrophysical Characterization of Porous Media », *Front. Water*, vol. 3, janv. 2022, doi: 10.3389/frwa.2021.800369.

[29] Y. Tai, J. Yang, X. Liu, et C. Xu, « MemNet: A Persistent Memory Network for Image Restoration ». arXiv, 7 août 2017. Consulté le: 11 mai 2023. [En ligne]. Disponible sur: http://arxiv.org/abs/1708.02209

[30] NVIDIA, « GoPro-OpenDataLab ». Consulté le: 19 août 2023. [En ligne]. Disponible sur:

https://seungjunnah.github.io/Datasets/gopro

[31] R. Jaesung, L. Haeyun, W. Jucheol, et C. Sunghyun, « Real-World Blur Dataset for Learning and Benchmarking Deblurring Algorithms ». 2020. Consulté le: 19 août 2023. [En ligne]. Disponible sur: http://cg.postech.ac.kr/research/realblur/

[32] Li *et al.*, « RESIDE-Standard ». 2019. Consulté le: 19 août 2023. [En ligne]. Disponible sur: https://sites.google.com/view/reside-dehaze-datasets/reside-standard

[33] A. Abdelrahman, L. Stephen, S. dBrown Michael, York University, et Microsoft Research, « Denoising Dataset for Smartphone Cameras ». 2020. Consulté le: 19 août 2023. [En ligne]. Disponible sur: https://www.eecs.yorku.ca/~kamel/sidd/dataset.php

[34] R. Dongwei, « datasets - OneDrive ». 2023. Consulté le: 19 août 2023. [En ligne]. Disponible                                                                                               sur: https://onedrive.live.com/?authkey=%21AIYIy8ZKL9kkmd4&id=66CE859AB42DFA2%2130 078&cid=066CE859AB42DFA2

[35] Wei, Chen and Wang, Wenjing and Yang, Wenhan and Liu, et Jiaying, « LOL Dataset - Machine Learning Datasets ». 2018. Consulté le: 19 août 2023. [En ligne]. Disponible sur: https://datasets.activeloop.ai/docs/ml/datasets/lol-dataset/

[36] B. Vladimir, P. Sylvain, C. Eric, et D. Fredo, « MIT-Adobe FiveK dataset:Learning Photographic Global Tonal Adjustment with a Database of Input / Output Image Pairs ». 2011. Consulté      le:      19      août      2023.      [En      ligne].      Disponible      sur: https://data.csail.mit.edu/graphics/fivek/

[37] Timofte *et al.*, « DIV2K Dataset ». 2018. Consulté le: 20 août 2023. [En ligne]. Disponible sur: https://data.vision.ee.ethz.ch/cvl/DIV2K/

[38] K. He, X. Zhang, S. Ren, et J. Sun, « Deep Residual Learning for Image Recognition ». arXiv, 10 décembre 2015. Consulté le: 22 août 2023. [En ligne]. Disponible sur: http://arxiv.org/abs/1512.03385

[39] N. Qian, « On the momentum term in gradient descent learning algorithms », *Neural Netw. Off. J. Int. Neural Netw. Soc.*, vol. 12, p. 145‑ 151, févr. 1999, doi: 10.1016/S0893-6080(98)00116-6.

[40] Z. Tu *et al.*, « MAXIM: Multi-Axis MLP for Image Processing ». arXiv, 1 avril 2022. Consulté le: 30 août 2023. [En ligne]. Disponible sur: http://arxiv.org/abs/2201.02973

[41] S.-H. Tsang, « Review: RDN — Residual Dense Network (Super Resolution) », Medium. Consulté le: 30 août 2023. [En ligne]. Disponible sur: